\documentclass[letterpaper, 10 pt, conference]{ieeeconf}  

\IEEEoverridecommandlockouts                              

\usepackage{graphics} 
\usepackage{epsfig} 
\usepackage{amsmath} 
\usepackage{amssymb}  
\usepackage{mathtools}

\usepackage[table]{xcolor}

\usepackage{xcolor,colortbl}
\definecolor{green2}{rgb}{0.1,0.5,0.1}
\definecolor{blue2}{rgb}{0.,0.75,0.75}

\newcommand\kevinupdate[1]{\textcolor{black}{#1}}
\newcommand\kevincomment[1]{\textcolor{black}{#1}}

\title{\LARGE \bf
Weakly-supervised DCNN for RGB-D Object Recognition in Real-World Applications Which Lack Large-scale Annotated Training Data}

\author{Li Sun$^{1}$ and Cheng Zhao$^{1}$ and Rustam Stolkin$^{1}$
\thanks{$^{1}$Extreme Robotics Lab, University of Birmingham, Birmingham, B15 2TT, UK.
        {\tt\small lisunsir@gmail.com@gmail.com}}%
}

\begin{document}

\maketitle
\thispagestyle{empty}
\pagestyle{empty}

\begin{abstract}

This paper addresses the problem of RGBD object recognition in real-world applications, where large amounts of annotated training data are typically unavailable. To overcome this problem, we propose a novel, weakly-supervised learning architecture (DCNN-GPC) which combines parametric models (a pair of Deep Convolutional Neural Networks (DCNN) for RGB and D modalities) with non-parametric models (Gaussian Process Classification). Our system is initially trained using a small amount of labeled data, and then automatically propagates labels to large-scale unlabeled data. We first run 3D-based objectness detection on RGBD videos to acquire many unlabeled object proposals, and then employ DCNN-GPC to label them. As a result, our multi-modal DCNN can be trained end-to-end using only a small amount of human annotation. Finally, our 3D-based objectness detection and multi-modal DCNN are integrated into a real-time detection and recognition pipeline. In our approach, bounding-box annotations are not required and boundary-aware detection is achieved. We also propose a novel way to pretrain a DCNN for the depth modality, by training on virtual depth images projected from CAD models. We pretrain our multi-modal DCNN on public 3D datasets, achieving performance comparable to state-of-the-art methods on Washington RGBS Dataset. We then finetune the network by further training on a small amount of annotated data from our novel dataset of industrial objects (nuclear waste simulants). \kevinupdate{Our weakly supervised approach has demonstrated to be highly effective in solving a novel RGBD object recognition application which lacks of human annotations. }
\end{abstract}

\section{INTRODUCTION}

DCNN-based approaches enable end-to-end inference, and can be highly parallelized to achieve computational efficiency. Unfortunately most DCNN methods rely on large-scale annotation of training data, which may be unavailable when trying to rapidly train such methods for new applications. Our work is motivated by the problem of training an RGBD object detection and recognition system for guiding robotic handling of hazardous nuclear waste, which may contain a vast array of different kinds of objects and materials, where massive acquisition and human-annotation of training data is not practical. 

\begin{figure}[thpb]
\centering
\includegraphics[width= 0.48\textwidth]{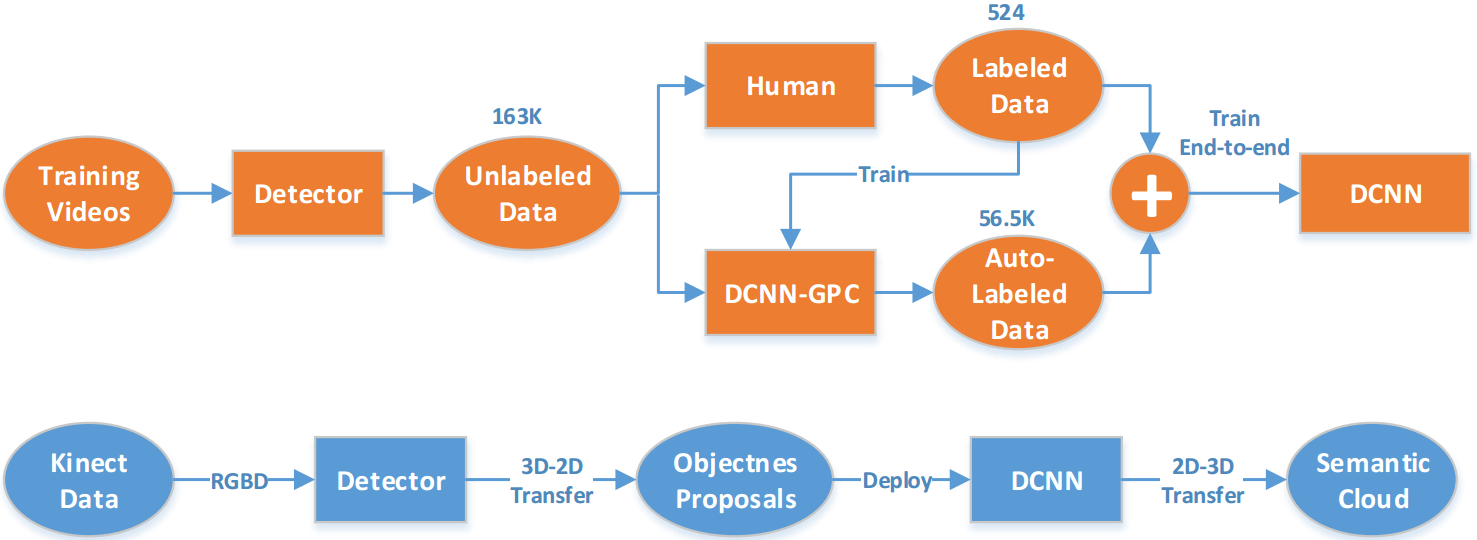}
\caption{\label{fig:weakly-supervised-learning}\kevinupdate{Flow chart of our proposed weakly-supervised DCNN method. Training is shown in orange and deployment in blue.}}
\end{figure}

To overcome this problem, we use a minimal amount of labeled data (0.3\%) to train a classifier which then automatically labels large-scale unlabeled data to enable end-to-end DCNN learning, i.e. weakly-supervised deep learning. For acquiring the unlabeled data, we deploy an unsupervised objectness detector on RGBD videos, Fig. \ref{fig:weakly-supervised-learning}.


\begin{figure}[thpb]
\centering
\includegraphics[width= 0.49\textwidth]{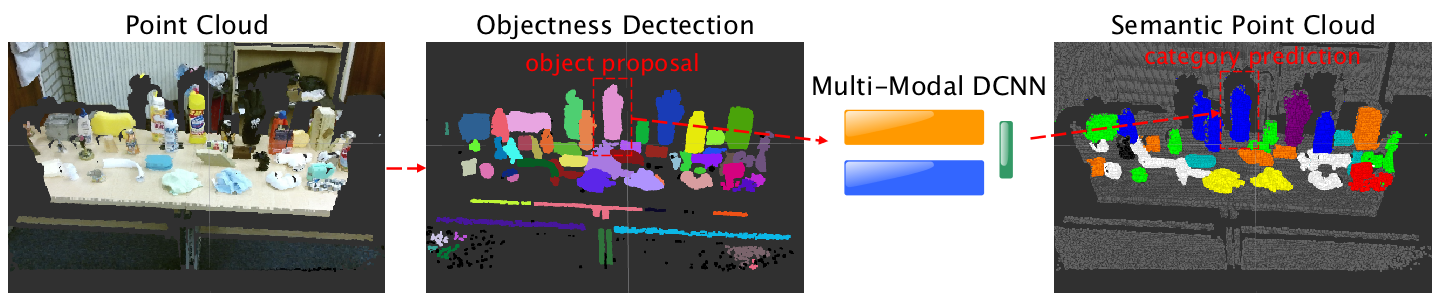}
\caption{\label{fig:detection-recognition}\kevinupdate{Detection and recognition pipeline of our system. RGBD point cloud (left) yields objectness proposals (middle). For each such proposal, the multi-modal DCNN performs category recognition. The pixel-wise recognition result is projected to obtain a 3D semantic cloud.}}
\end{figure}

The main contributions of this paper are as follows:
\begin{itemize}
\item Previous RGBD object recognition methods have predominantly been fully-supervised, making them unsuitable for rapid deployment in new applications. In contrast, we propose a weakly supervised method, based on Gaussian Process Classification (GPC) combined with DCNN deep learning. Unlike previous work, our method does not require bounding box object annotation, and uses very little manually-labeled data (0.3\%).

\item Our approach learns directly from raw depth images, in contrast to previous work which relies on extracting low-level features or color-mapping. We use a new approach to pretrain the depth DCNN by using many automatically generated synthetic depth images.

\item We introduce a new industrial dataset\footnote{The video demo, dataset, and our implementation are available online: https://sites.google.com/site/romansbirmingham/}, comprising RGBD videos of realistic nuclear waste-like objects. A real-time detection and recognition system is implemented and significantly outperforms a fully-supervised method i.e R-CNN on this real-world data. 
\end{itemize}

\section{Related Work}

\subsection{Objectness and object detection}\label{sec:review_detection}
Most object detection literature addresses only 2D RGB images, e.g. \cite{uijlings2013selective,arbelaez2014multiscale}. Region-based CNN (R-CNN) \cite{rcnn} frameworks comprise: objectness detection, then pretrained-CNN-based feature extraction, followed by SVM classifiers for object category recognition. More recent work, \cite{fast-rcnn,yolo,ssd}, achieves greater speed by using DCNNs, in which both detection and recognition can be learned jointly and deployed in a single shot. However, DCNNs depend on large-scale human-annotated training data which is often unavailable in real-world applications. Furthermore, these methods are based on bounding-box detection, and cannot achieve boundary-aware detection.

Comparatively little literature has addressed the use of 3D data, which can greatly facilitate objectness detection by providing more salient boundaries between foreground objects and background regions. \kevinupdate{Some research simply adapts 2D mechanism to RGBD \cite{gupta2014learning} without the consideration of the real 3D distance metric.} \cite{song2014sliding} detects 3D objects in a point cloud by applying a cuboid-shaped sliding window. \cite{song2016deep} extended the region proposal networks of \cite{multibox} to achieve faster object detection than sliding-shape approaches. However, these methods typically generate thousands of objectness proposals for each image, making subsequent object category recognition difficult to do in real-time.

Alternatively, unsupervised 3D segmentation (clustering) \cite{asif2016unsupervised} can be used for RGBD objectness detection, and can also achieve boundary-aware detection. Such methods engender a trade-off between segmentation accuracy and speed. In our approach, we simplify the 3D clustering connectability, using only three cues, to enable real-time performance while still achieving boundary-aware detection.





\subsection{RGBD object recognition}\label{sec:review_recognition}

Multimodal DCNNs \cite{schwarz2015rgb,eitel2015multimodal,cheng20153dv,cheng2015iccv,cheng2016semisupervised} are now widely used in RGB-D object recognition. These multimodal architectures comprise two nets (for RGB and D modalities) which are fused in the last fully-connected layers and trained jointly. These methods pretrained both RGB and D DCNNs on ImageNet, since no large-scale labeled depth dataset was available for pretraining. Unfortunately, network parameters pretrained on RGB images (i.e. ImageNet) do not compliment raw depth data well. Most methods transfer depth modality to RGB modality through color-mapping \cite{schwarz2015rgb,eitel2015multimodal, hoffman2016cross}, or low-level features \cite{cheng2015iccv,cheng2016semisupervised,wang2016differential}, to fit into a DCNN pretrained on RGB data (ImageNet). These methods need extra computation for color-mapping and feature extraction, and the raw depth data is not fully leveraged. In contrast to previous work, our DCNN is directly trained on raw depth maps. No costly data conversions (from depth to RGB) are required, and depth information is fully exploited.


\subsection{Weakly-supervised deep learning}\label{sec:review_wdl}
Following the success of highly data-driven DCNN methods, the problem of reducing annotation effort has attracted increasing attention. Cheng, et al. \cite{cheng2016semisupervised} proposed semi-supervised learning approaches for RGBD object recognition, in which co-training methods are used to incrementally label the unlabeled data. Papandreou, et al. \cite{weakly-fcn} proposed a weakly supervised DCNN to learn pixel-wise semantic segmentation from bounding-box annotations. In their method, dense CRF is used to obtain segmentation estimations for training DCNN.

Barnes, et al. \cite{barnes2016find} utilize the temporal correlation in driving videos to learn path proposals for autonomous driving. In their method, the path in future frames are projected to the current frame through vehicle odometry and annotated as ground truth for learning. Zeng, et al. \cite{zeng2016multi} proposed a self-supervised approach to learn fully-constitutional networks for object segmentation in the Amazon picking challenge.


The key step in semi-supervised or weakly-supervised learning in object recognition is to model the predictive probability. In \cite{cheng2016semisupervised}, the DCNN trained by labeled examples is used as the classifiers in co-training. However, small-scale training examples open up the possibility of over-fitting, and as a consequence, a good predictive probability cannot be guaranteed. In contrast, we adapt non-parametric GP classification with a fusion of multi-modal kernels, which is more robust to the scale of training data. We reduce the required label percentage from 5\% \cite{cheng2016semisupervised} to 0.3\% (at the same fps rate). \kevincomment{More importantly, their research \cite{cheng2016semisupervised} focus on recognition which trained by bounding-box annotations, whereas our research is weakly-supervised with integration of objectness detection, learned by detected objectness proposals.}


\subsection{Discussion}\label{sec:discussion}
Compared to 2D-based detection methods \cite{uijlings2013selective,multibox,fast-rcnn,yolo,ssd,multibox}, 3D-segmentation-based detectors can reduce the number of object proposals from thousands to less than a hundred per image. More importantly, boundary-aware detection can be obtained. In  our approach, we propose a 3D-segmentation method which is multi-cue, but more efficient than \cite{asif2016unsupervised}, for real-time objectness detection in RGBD data.

Multi-modal DCNNs achieve state-of-the-art performance in RGBD object recognition. However, how to fully leverage the depth modality remains a problem. Recent work \cite{eitel2015multimodal,cheng2015iccv,cheng2016semisupervised,wang2016differential} assumes that raw depth images cannot be directly used to train a DCNN, because no large-scale depth dataset is available for pretraining. In contrast, we show how raw depth data can be fully leveraged, by using 3D CAD models (e.g. Model-Net) to generate large numbers of automatically annotated depth images for pretraining. As a consequence, color-mapping methods and low-level depth features are not required in our approach.

Most DCNN-based detection and recognition methods are fully supervised, trained by massive annotated datasets. In contrast, weakly-supervised deep learning has, so far, only achieved success in very few applications, including path planing \cite{barnes2016find}, and Amazon picking challenge\cite{zeng2016multi}. In contrast, this paper shows how weakly supervised deep learning can achieve very strong performance on RGBD object detection and recognition, at real-time frame rates, on real-world industrial image data, for which only a tiny amount (0.3\%) of labeled data is available for training.


\section{Preliminaries}

\subsection{Gaussian Process Classification (GPC)}

Unlike popular classifiers e.g. SVM, GPC is fully Bayes-based and extendable to multi-class cases. Our use of GPC follows \cite{gpml}. Given a GPC problem: training instances $X$, training labels $y$, testing instance $x_*$, testing label $y_*$, and  latent variables for training and testing instances $f$ and $f_*$, the GPC infers the conditional predictive probability of testing instance's label $y_*$ given $X$ and $y$:

\begin{equation}
\small
\begin{split}
P(y_*|x_{*},X,y) = \iint P(y_*|f_*)p(f_*|f,x_*,X)p(f|X,Y)~df_{*}df.
\end{split}\label{eq:confidional_prob}
\end{equation}
where $P(y_*|f_*)$ is prohibit distribution for binary classification, and $p(f_*|f,x_*,X)$ is a standard noise-free regression. The key problem of GPC is to estimate the posterior $p(f|X,y)$:

\begin{equation}
\small
p(f|X,y)=\frac{P(f|X)P(y|f)}{\int P(f|X)P(y|f) df}\label{eq:posterior}
\end{equation}
where $P(f|X)$ is the prior, $P(y|f)$ is the likelihood. Here, the prior Gaussian, whereas the likelihood is non-Gaussian, which makes Eq. \ref{eq:posterior} analytically intractable. Researchers proposed different ways \cite{gpml} to solve this non-conjunction problems, including Laplace Approximation, Expectation Propagation, etc.


\section{Methodology}

\begin{figure}[t]
\centering
\includegraphics[width= 0.5\textwidth]{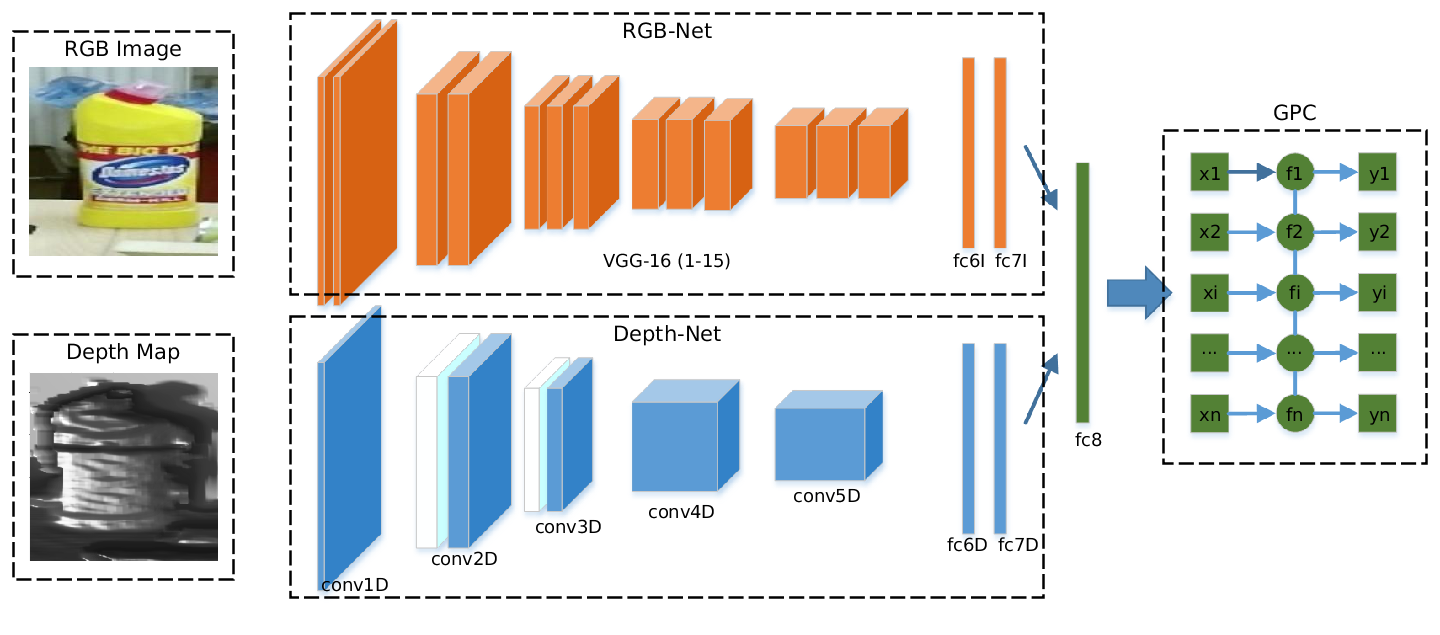}~~
\caption{\label{fig:cnns-gpc}The architecture of proposed multi-modal DCNN-GPC. The inputs of the DCNNs are the raw RGB image and depth map of the object proposal. Our architecture consists of three components: RGB-Net (shown in yellow)  Depth-Net (shown in Blue) and non-parametric GPC (shown in Green). The parameters of Depth-Net are detailed in Table 1.}
\end{figure}

\begin{table}[thpb]
\scriptsize
\centering
\caption{\label{tab:architecture} The Neural Network Architecture of our Depth-Net.}
\begin{tabular}{|p{1.2cm}|p{1.cm}|p{1.3cm}|p{2.8cm}|}
\hline
layer name & filter size & no. of output & other parameters\\
\hline
conv1D & 5 $\times$ 5 & 128 & stride=2, pad=2, group=1 \\
\hline
pool1D & 3 $\times$ 3 & -- & type=max, stride=2, pad=1 \\
\hline
norm1D & 5 $\times$ 5 & -- & alpha=5$\times 10^{-4}$ beta=0.75 \\
\hline
conv2D & 5 $\times$ 5 & 256 & stride=1, pad=2 , group=1 \\
\hline
pool2D & 3 $\times$ 3 & -- & type=max, stride=2, pad=2 \\
\hline
norm2D & 5 $\times$ 5 & -- & alpha=5$\times 10^{-4}$ beta=0.75 \\
\hline
conv3D & 3 $\times$ 3 & 384 & stride=1, pad=1, group=2 \\
\hline
pool3D & 3 $\times$ 3 & -- & type=max, stride=2, pad=2 \\
\hline
conv4D & 3 $\times$ 3 & 512 & stride=1, pad=1, group=1\\
\hline
conv5D & 3 $\times$ 3 & 512 & stride=1, pad=1, group=1\\
\hline
pool5D & 3 $\times$ 3 & -- & type=max, stride=2, pad=2 \\
\hline
fc6D & --  & 4096 & dropout = 0.5 \\
\hline
fc7D & --  & 4096 & dropout = 0.5 \\
\hline
fc8D & --  & c & -- \\
\hline
\end{tabular}
\end{table}

Our proposed pipeline has three steps: (1) a real-time 3D-based object detection approach is proposed to generate high-quality objectness proposals in RGBD video stream; (2) DCNN-GPC is proposed to propagate small-scale labeled data to moderate-scale in order to train the multi-modal DCNN end-to-end; (3) a real-time detection and recognition system is integrated.

\subsection{Real-time 3D Object Detection}
Our object detection approach is 3D-based and unsupervised, employing point cloud segmentation to obtain salient objectness (regions) proposals. We first detect large planes (using RANSAC) in point clouds and remove them, as we are interested in table-top or ground-top objects. Inspired by the multi-cues idea of \cite{asif2016unsupervised}, we propose a more efficient conditional clustering based on color, shape and spatial cues to acquire objectness proposals. Given two voxels $p_1$ and $p_2$, the connectability between them $\mathcal{C}(p_1,p_2)$ is defined by distance connectability $\mathcal{C}_d(p_1,p_2)$, color connectability $\mathcal{C}_c(p_1,p_2)$ and shape connectability $\mathcal{C}_s(p_1,p_2)$:

\begin{equation}
\small
\begin{split}
\mathcal{C}_{s}(p_1,p_2) = \begin{dcases}
1, & \text{if } \frac{\textbf{n}_{p1} \cdot \textbf{n}_{p2}}{\parallel \textbf{n}_{p1}\parallel \parallel \textbf{n}_{p2}\parallel} < \sigma_s ~~~~~~~~~~~~~~~~~~~~~~\\
0, & \text{otherwise}~~~~~~~~~~~~
\end{dcases}
\\
\mathcal{C}_{c}(p_1,p_2) = \begin{dcases}
1, & \text{if } \parallel I_{p1} - I_{p2} \parallel < \sigma_c ~~~~~~~~~~~~~~~~~~~~~~~~~\\
0, & \text{otherwise}
\end{dcases}
\\
\mathcal{C}_{d}(p_1,p_2) = \begin{dcases}
1, & \text{if } \parallel p_1 - p2_2 \parallel < \sigma_d ~~~~~~~~~~~~~~~~~~~~~~~~~\\
0, & \text{otherwise}
\end{dcases}
\\
\mathcal{C}(p_1,p_2) = \mathcal{C}_{d}(p_1,p_2) \cap (\mathcal{C}_{s}(p_1,p_2) \cup \mathcal{C}_{c}(p_1,p_2))~~~~~~~~~~~~~~
\end{split}
\end{equation}

where $\textbf{n}_{p1},\textbf{n}_{p2}$ are the surface normals, ${I}_{p1},{I}_{p2}$ refer to the intensity values of $p_1,p_2$, and $\sigma_d$,   $\sigma_c$,  $\sigma_s$ are the connectability thresholds. The neighboring voxels will be clustered iteratively through this connectability criteria until all clusters become constant.  Parameter values $\sigma_d$,   $\sigma_c$,  $\sigma_s$ are set as 2.0CM, 8.0 and 10$^{\circ}$, perform well for our application.

Given 3D objectness proposals detected in 3D world coordinates, each point in the proposal $p(x_w,y_w,z_w)$ can be back-projected to its 2D image coordinates $(u,v)$ and it depth $d$: 
\begin{equation}\label{eq:2d3d}
d~ 
\begin{bmatrix}
    u\\
    v\\
    1
\end{bmatrix} =
C~
\begin{bmatrix}
    R~t\\
    0~1\\
\end{bmatrix}
~
\begin{bmatrix}
    x_w\\
    y_w\\
    z_w
\end{bmatrix}
\end{equation}

where $C$ is the camera intrinsic matrix, and $R$ and $t$ are the rotation matrix and transformation vector respectively. In this case, a 2D bounding box with boundary-aware segmentation can be formed for each 3D objectness proposal, which is used for learning in the following section.

\subsection{Weakly-Supervised Multi-Modal DCNN for RGBD Object Recognition}
Similar to popular DCNN-based methods \cite{schwarz2015rgb,eitel2015multimodal,cheng20153dv,cheng2015iccv,cheng2016semisupervised}, our recognition DCNN is also of multi-modal architecture (RGB-Net and Depth-Net for RGB and depth modalities respectively). However, we propose a different DCNN architecture and propose a novel weakly-supervised way to train it, comprising three stages. First, the DCNNs are pretrained on public large-scale datasets (ImageNet for the RGB-net and Model-Net for the Depth-net). Second, the DCNN-GPC is trained and then employed to classify large-scale unlabeled objectness proposals according to predictive probabilities of the GPC. Third, the multi-modal DCNN, used in the DCNN-GPC, is fine-tuned jointly end-to-end using moderate-scale automatically labeled RGBD data.

\subsubsection{Network Architecture}
in contrast to Caffe-Net (used in \cite{schwarz2015rgb,eitel2015multimodal,cheng20153dv,cheng2015iccv,cheng2016semisupervised}), we use a deeper architecture for RGB-modality recognition. That is, the VGG 16 layers' architecture is used for our RGB-Net with the removal of softmax layer. For Depth-Net, we devised an 8 layers DCNN. Compared to the widely-used Caffe-Net \cite{caffe-net}, our Depth-Net adapts smaller filter sizes and larger numbers of filters. The Local Response Normalization (LRN) layers \cite{alex-net} are applied to the first two convolutional layers' features in order to capture the local 3D shape of the object from the relative range difference. All convolutional layers and fully connected layers are initialized by Xavier initialization \cite{xavier}. The parameters of this architecture are set according to experimental experience (shown in Table \ref{tab:architecture}). .

\subsubsection{Pretraining of Multi-modal DCNN}\label{sec:pretraining}

To train large-capacity DCNNs without over-fitting, pretraining is necessary. Our RGB-Net is pretrained on ImageNet \cite{image-net}. In well-known methods \cite{schwarz2015rgb,eitel2015multimodal,cheng20153dv,cheng2015iccv}, the DCNN for depth-modality recognition is also pretrained on ImageNet, requiring color-mapping or low-level features to transform raw depth data into the RGB domain.  In contrast, our proposed Depth-Net is pretrained on the Model-Net dataset \cite{model-net} from scratch, by projecting many synthetic depth maps. As a result, no extra pre-processing (colormaping or low-level features) is needed. 

\begin{figure}[thpb]
\centering
\includegraphics[width= 0.17\textwidth]{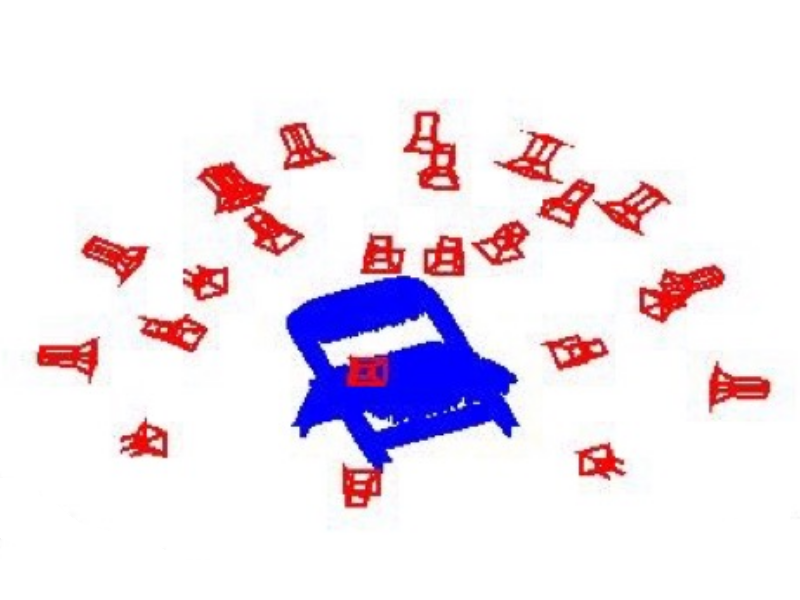}
\includegraphics[width= 0.17\textwidth]{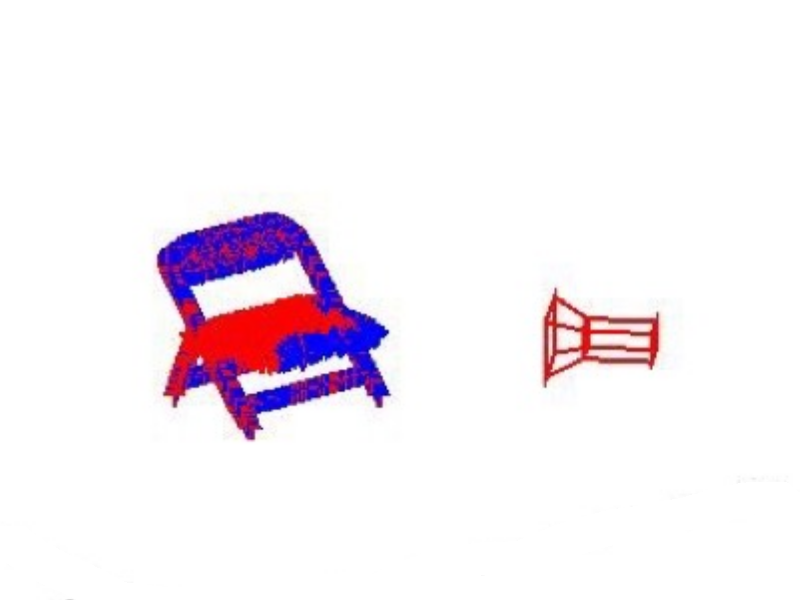}
\includegraphics[width= 0.13\textwidth]{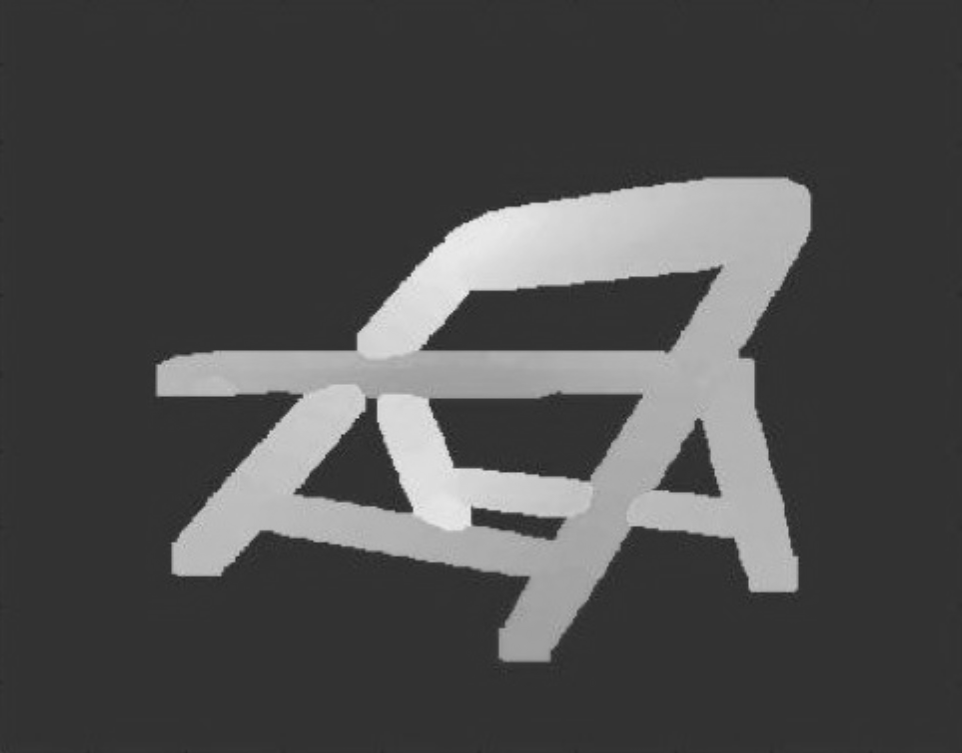}
\caption{\label{fig:virtual_camera} Fig-a, example camera poses. Fig-b, Hidden Points Removal is applied on 3D points to remove occluded points from camera view.  Fig-c, the depth map is captured.}
\end{figure}

In our approach, we use 40 class subsets of Model-Net (in total 9.8K models) for training. For each 3D model, we sample 4$\times 10^{4}$ points uniformly on the object surface and apply white noise on those 3D points. Then for the point cloud of each object, we generate 30 camera poses, distributed on a hemisphere, and capture depth maps from each camera pose\footnote{More specifically, the poses of virtual cameras are obtained by discretizing Euler angles: roll are 270$^{\circ}$, 240$^{\circ}$, 210$^{\circ}$, pitch is fixed at 0$^{\circ}$, and yaw are ranging from 0$^{\circ}$ to 360$^{\circ}$ with interval of 36$^{\circ}$.}. After 6DOF camera poses are obtained, for each camera the inverse transform is applied to the original point cloud to transfer 3D points from world to camera coordinate systems. Hidden points removal \cite{HPR} is then applied on the 3D points, and a depth image is generated by projecting visible 3D points to the image plane via Eq. \ref{eq:2d3d}.\footnote{In our implementation, the optical center of our virtual camera is set as (250, 250), and focal length is 500. Consequently, a depth map of 500$\times$500 resolution is obtained. We resize the depth maps to (224,224) for training.}. The overall process is shown in Fig. \ref{fig:virtual_camera}. Finally, 290K depth maps are obtained from training models.

\subsubsection{Label Propagation through DCNN-GPC}
As shown in Fig. \ref{fig:cnns-gpc}, DCNN-GPC incorporates pretrained DCNNs (with softmax layer removed) and non-parametric GP classification. That is, the outputs of DCNNs (i.e. $fc7_I$ $\in$ $\mathbb{R}^{4096}$ and $fc7_D$ $\in$ $\mathbb{R}^{4096}$) are concatenated as the input $X$ ($\in \mathbb{R}^{8192}$) in GPC. \kevinupdate{In this stage, we label a small number of detected objectness proposals and train GPC on them.}

More specifically, the training data of our proposed method are unlabeled RGBD videos. We capture the videos in a controlled environment, i.e. only one object category is recorded in each video. By deploying the proposed detector to the training videos, for each category, a large-scale unlabeled objectness proposals set $\{\mathcal{S}^{U}\}$ can be obtained. We then manually label a very small sub-set $\{\mathcal{S}_{m}^{L}\}$ of objectness proposals and train GPC on $\{\mathcal{S}_{m}^{L}\}$.


In our approach, the binary GPC is adapted, and can easily be extended to multi-class GPC if the environment is not controllable. More specifically, the prior $P(f|X)$ is modeled as Gaussian $\mathcal{N}(0,K)$, where $K$ is the covariance matrix of all training examples $X$. In order to interpret features from different modalities, we treat the kernel as the product of kernels of different data domains:

\begin{equation}
k(x,x')= k_I(x_{I},x_{I}') * k_D(x_{D},x_{D}')
\end{equation}

where $x^{(,)}_{I}$ is first 4096 dimensional features produced by $fc7I$ of RGB-net and $x^{(')}_{D}$ refers to the last 4096 dimensional features produced by $fc7D$ of Depth-net, $x=[x^{(')}_{I},x^{(')}_{D}]$. In our approach, RBF kernel is used for $k_1$ and $k_2$:

\begin{equation}
\centering
k_{RBF}(x,x') = \alpha^{2} \exp^{ - \frac{\parallel x - x' \parallel}{2\beta^{2}}} ,
\end{equation}

The scale parameter $\alpha$ and deviation parameter $\beta$ are hyper-parameters of the kernel. Consequently, $k$ has four hyper-parameters.

In order to solve this non-conjunction problem in the posterior estimation of the GPC, Expectation Propagation (EP) \cite{gpml} is used in our approach. Eq. \ref{eq:posterior} can be rewritten provided the training examples are i.i.d.:

\begin{equation}
\footnotesize
p(f|X,y)=\frac{P(f|X)P(y|f)}{\int P(f|X)P(y|f) df}
= \frac{P(f|X)\prod P(y_i|f_i)}{\int P(f|X)\prod P(y_i|f_i) df}
\label{eq:posterior2}
\end{equation}

In EP, each local likelihood $P(y_i|f_i)$, is treated as a non-normalized Gaussian, thereby estimating the global likelihood as a multivariate Gaussian (with a multiplication of normalization terms). Moreover, the hyper-parameters mentioned above are optimized through maximizing the log marginal likelihood (the log of the denominator of Eq. \ref{eq:posterior2}). As investigated in our previous research \cite{sun2016icra}, this step is of significant importance as the predictive probability can be well-spread after hyper-parameter optimization. More details about EP and hyper-parameters estimation can be found in \cite{gpml}.

After GPC is trained and hyper parameters optimized, we employ GPC to propagate labels to large-scale unlabeled dataset $\{\mathcal{S}^{U}\}$.  We model the prediction confidence as the predictive probability of GPC:

\begin{equation}
confidence = P(y_*|x_{*},X,y)\label{eq:confidional_prob}
\end{equation}

We set a confidence interval $\in [\tau, 1]$ and assign an object label to those examples whose prediction confidence lies in this interval, yielding a moderate-scale of labeled data $\{\mathcal{L}_{GP}\}$.


\subsubsection{End-to-End Training using DCNN-GPC Labeled Data}
Having large-scale unlabeled data automatically labeled by DCNN-GPC, sufficient training examples i.e. $\{\mathcal{S}_{m}^{L}\}$  and $\{\mathcal{S}_{GP}^{L}\}$ are obtained to train RGB-Net and Depth-Net from-end-to-end. At this stage, we replace GPC with a softmax loss layer, connected with fully-connected layer fc8. We extend the conventional multi-modal softmax loss (i.e. negative log likelihood) \cite{eitel2015multimodal} to the weakly-supervised case:

\begin{equation}
\footnotesize
\begin{split}
loss = - \sum_{i \in \mathcal{S}_{m}^{L}} log \mathcal{L}(softmax(f^{fc8}([O_i^{fc7_I}, O_i^{fc7_D}], \theta^{fc8}), y_i) ~~~~~~~~~~~~~~\\
-  \eta \sum_{j \in \mathcal{S}_{GP}^{L}} log \mathcal{L}(softmax(f^{fc8}([O_j^{fc7_I}, O_j^{fc7_D}], \theta^{fc8}), y_j)~~~~~~~~~~~~~~
\end{split}
\end{equation}

where $O_*^{fc7_I(D)}$ is the output of $fc7_I(D)$, $\mathcal{L}$ is the likelihood function, $\theta$ is the parameters of $fc8$ and $y_{i(j)}$ refers to the training label. $\eta \in [0,1]$ is the penalty factor of DCNN-GPC automatically labeled training data, set according to the automatic annotation quality. In our implementation, the loss of DCNN-GPC-labeled examples are treated equivalently to that of manually-labeled examples ($\eta$=1) as our DCNN-GPC yields satisfactory annotations. 

\section{Experiments}
We report three sets of experimental results. Since we use 3D CAD models to pretrain our Depth-Net, we firstly evaluate our Depth-Net's performance using the Model-Net 3D model classification challenge \cite{model-net} (section \ref{sec:exp_modelnet}). In Section \ref{sec:exp_washington}, we verify the adaptation of our DCNN (pretrained on CAD models) using the Washington RGBD object recognition benchmark \cite{washington}. Finally, we evaluate our proposed weakly-supervised deep learning approach, and our integrated detection and recognition system, using a novel real-world application, using our new dataset of industrial objects (nuclear waste simulants) in section \ref{sec:exp_romans}. 

\subsection{Model-Net 3D Model Classification Challenge}\label{sec:exp_modelnet}

In this experiment, the 40-class subset of the Princeton Model-Net dataset\footnote{Dataset available at: http://modelnet.cs.princeton.edu/} is used. There are 12.4K 3D CAD models in total (9.8K for training, 2.4K for validation). Following the procedure illustrated in Section \ref{sec:pretraining}, 290K depth maps are obtained from training models. A mini-batch of 128 is used for SGD learning. The learning rate is set as 0.01 with a reduction of 10 times every 10K iterations. Training converges after 30K iterations. The momentum is fixed as 0.9 and weight decay is 5$\times 10^{-4}$.

\begin{table}[thpb]
\caption{\label{tab:model-net}. \scriptsize Comparison with top-ranking methods in Model-Net (40-class). }
\centering
\scriptsize
\begin{tabular}{|p{2.8cm}|p{2.7cm}|p{2cm}|}
\hline
Method & Classification Accuracy\\
\hline
Geometry Image\cite{sinha2016deep} & 83.9\% \\
\hline
Set-convolution\cite{ravanbakhsh2016deep} & 90\%  \\
\hline
VRN Ensemble\cite{brock2016generative} & \textbf{95.54\%} \\
\hline
FusionNet\cite{HegdeZ16} & 90.8\%  \\
\hline
Pairwise\cite{johns2016pairwise} & 90.7\% \\
\hline
MVCNN\cite{su2015multi} & 90.1\% \\
\hline
3DShapeNets\cite{model-net} & 77.0\% \\
\hline
\hline
Ours & \textbf{92.7\%} \\
\hline
\end{tabular}
\end{table}

Since our goal is to utilize Model-Net to pretrain our DCNN (not to optimise 3D model classification to maximise performance on the Model-Net challenge), in our approach, we minimize the average negative log likelihood of all 2.5D views:

\begin{equation}
\footnotesize
loss = - log \sum_{i=1}^{N_model} \sum_{j=1}^{N_view} \mathcal{L}(softmax(O_{ij}^{fc_8^D}), y_{ij})
\end{equation}

In the testing phase, the model classification prediction can be obtained by simply averaging the predictions of all fields of view. Though no special fusion method is used for model label prediction, our approach still delivers superior results to all compared state-of-the-art approaches except \cite{brock2016generative} (Table \ref{tab:model-net}). \kevinupdate{In \cite{brock2016generative}, the volumetric representation is used, thereby eliminating the occlusions in 2.5D views. In contrast, our objective is to pretrain our DCNN and transfer the model to Kinect data, hence 2.5D depth map is used. Our approach achieves the best result out of all methods using 2.5D representations. }

\subsection{Washington RGBD Object Recognition Dataset}\label{sec:exp_washington}
The Washington RGBD object dataset\footnote{https://rgbd-dataset.cs.washington.edu/} comprises 300 objects organized in 51 categories.  In this experiment, the evaluation set (a subset of every 5 frames, in total of 41,877 RGBD images) reported in Lai et. al \cite{lai2011a} are used. We follow the original train/validation splits.

\begin{table}[thpb]
\centering
\scriptsize
\caption{\label{tab:washington}\scriptsize Comparison with top-ranking methods in Washington RGBD dataset.}
\begin{tabular}{|p{2.8cm}|p{1.3cm}|p{1.3cm}|p{1.3cm}|}
\hline
Method & RGB & Depth & RGB-D \\
\hline
CNN-RNN\cite{socher2012convolutional} & 80.8 $\pm$ 4.2 & 78.9 $\pm$ 3.8 & 86.8 $\pm$ 3.3  \\
\hline
Upgraded HMP\cite{bo2013unsupervised} & 82.4 $\pm$ 3.1 & 81.2 $\pm$ 2.3 & 87.5 $\pm$ 2.9  \\
\hline
Multi-Modal\cite{eitel2015multimodal} & 84.1 $\pm$ 2.7 & 83.8 $\pm$ 2.7 & 91.3 $\pm$ 1.4  \\
\hline
QASM\cite{cheng2015query} & -- & -- & \textbf{92.7 $\pm$ 1.0}  \\
\hline
Hypercube Pyramid\cite{zaki2016convolutional} & 87.6 $\pm$ 2.2 & 85.0 $\pm$ 2.1 & 91.1 $\pm$ 1.4 \\
\hline
Semi-Supervised \cite{cheng2016semisupervised} & 85.5 $\pm$ 2.0 & 82.6 $\pm$ 2.3 & 89.2 $\pm$ 1.3 \\
\hline
\hline
Ours & \textbf{88.4 $\pm$2.1 } & \textbf{80.3$\pm$ 2.7} & \textbf{91.8 $\pm$ 1.1 }\\
\hline
\end{tabular}
\end{table}

In this experiment, our multi-modal DCNN is trained in three steps. For initialization, the weights of RGB-Net and Depth-Net are pretrained on ImageNet and ModelNet datasets respectively. In the first step, we freeze the RGB-Net layers and fusion layers (disable $softmax_{fusion}$), and finetune the Depth-Net with mini-batch of 64, fixed learning rate $10^{-2}$ and weight decay $5\times 10^{-4}$. This training converges after 20K iterations. In the second step, we finetune both RGB-Net and Depth-Net (still disable $softmax_{fusion}$) with mini-batch of 32 and fixed learning rate $10^{-3}$ for another 10K iterations. Finally, similar to \cite{eitel2015multimodal}, we freeze the RGB-Net and Depth-Net layers and train the fusion layer. A mini-batch of 32, and fixed learning rate $5\times 10^{-4}$ are used, and training converges quickly after 5K iterations. Compared to other DCNN methods \cite{schwarz2015rgb,eitel2015multimodal,cheng20153dv,cheng2015iccv,cheng2016semisupervised}, we use a deeper architecture for the RGB modality, thereby achieving better RGB recognition performance. Moreover, unlike other methods, our Depth-Net uses raw depth data for training, i.e. real end-to-end learning between raw sensor data and the learning objective. Inference of depth modality is more straight-forward as no extra computation (color mapping or low-level features) are required. Moreover, our multi-modal DCNN is pretrained on different types of data, and more distinctive classification views are trained for different modalities. While performance of our depth-Net is slightly lower than the state-of-the-art, substantial improvement is then obtained by fusing multi-modal DCNNs.

As shown in Table \ref{tab:washington}, our DCNN achieves an average of 91.8\% recognition accuracy among 51 categories of objects, which outperforms all compared state-of-the-art approaches except \cite{cheng2015query}. \kevinupdate{In \cite{cheng2015query}, they first use an off-the-shelf recognition method \cite{socher2012convolutional} to generate top ranking candidate categories. They further propose a similarity metric between testing and training objects and make the prediction base on a voting of multiple nearest training objects. Hence, this is a higher-level ensemble method rather than a recognition method. In other words, our approach could also be significantly improved, in principle, by using their ensemble mechanism. Our approach achieves state-of-the-art performance among all DCNN-based methods. Most importantly, this result demonstrates a good adaptation of our pretrained Depth-Net from virtual data to real-world RGBD data.}


\subsection{Industrial Object Recognition}\label{sec:exp_romans}

\begin{table*}[t]
\centering
\scriptsize
\caption{\label{tab:result} \scriptsize Statistics of our dataset, training examples, and quantitative results of our proposed detection/recognition system. Detection precision rate, recall rate and f-score of each category are given. T.E stands for training examples, inst.w. for instance-wise and pix.w. for pixel-wise.}
\begin{tabular}{|p{2.8cm}|p{1.cm}|p{.8cm}|p{.8cm}|p{.8cm}|p{.8cm}|p{1.0cm}|p{1.0cm}|p{1.0cm}|p{1.cm}|p{1.1cm}|p{1.1cm}|}
\hline
Category        & \cellcolor{blue!80}bottles & \cellcolor{green!90}cans   & \cellcolor{red!90}chains & \cellcolor{yellow!90}cloth  & \cellcolor{purple!90}gloves & \cellcolor{black!75}metal obj. & \cellcolor{white!90}pipe join. & \cellcolor{green2!90}plas. pipe & \cellcolor{blue2!90}sponges & \cellcolor{orange!90}wood bloc. & overall/ave. \\
\hline
Instance Amount & 28 / 12     & 22 / 15    & 8  / 3   & 6 / 3   & 16 / 5    & 22 / 10         & 9 / 5          & 10  / 4        & 12  / 6   &  14 / 7        & 147 / 70\\
\hline
Videos          & 4       & 2      & 2      & 4      & 4      & 4             & 2            & 2           & 2       & 3           & 23 \\
\hline
Unlabelled T.E. & 20.5K   & 32.5K  & 18.3K  & 13.3K   & 8.6K   & 22.1K         & 21.9K        & 8.0K        & 9.0K    &  14.0K      & 163K\\
\hline
Labelled T.E.   & 48      & 56     & 26     & 45     & 35     & 48            & 28           & 20          & 32      & 32          & 524 \\
\hline
GP Labelled T.E.& 11436   & 15525  & 2322   & 4606   & 5298   & 6101          & 2287         & 1037         & 3223    & 4734       & 56.5K\\
\hline
\hline
Precise of R-CNN (inst.w.) &   68.10\%  &    72.57\%  &   69.77\%  &   62.26\%  &   48.94\%  &   60.00\%  &   44.58\%  &   72.22\%  &   62.26\%  &   67.86\%  & 64.63\% \\
\hline
Recall of R-CNN (inst.w.)  & 53.02\%  &   70.95\%  &   78.95\%  &   70.21\%  &   41.82\%  &   50.85\%  &   45.17\%  &   46.43\%  &   53.52\%  &   16.96\%  &  52.30\% \\
\hline
F-Score of R-CNN (inst.w.) & 59.62\%  &   71.75\%  &   74.07\%  &   66.00\%  &   45.10\%  &   55.05\%  &   45.02\%  &   56.52\%  &   57.57\%  &   27.14\%  &  57.81\% \\

\hline
Precise of Ours (inst.w.) &  89.19\%  &  81.82\%  &  79.17\%  &  93.33\%  &  68.25\%  &  75.00\%  &  66.67\%  &  63.16\%  &  92.45\% &   87.84\% &  \textbf{80.85\%}\\
\hline
Recall of Ours (inst.w.) & 83.19\%  &  91.84\%  &  95.00\%  &  80.00\%  &  91.49\%  &  64.04\%  &  90.20\%  &  50.00\%   & 87.50\%  &  87.84\% &  \textbf{83.53\%}\\
\hline
F-Score of Ours (inst.w.) & 86.09\%  &  86.54\%   &  86.36\%  &   86.15\%  &  78.18\%  &  69.09\%  &    76.67\%  &    55.81\%  &    89.91\%  &   87.84\% &  \textbf{82.17\%}  \\
\hline
\hline
Precise of R-CNN (pix.w.)  &  66.75\%  &   63.55\%  &  68.09\% &  58.03\%  &   55.24\%  &  45.35\%  &   57.30\%  &  43.81\%   &  55.09\%  &   59.43\%  &   59.46\% \\
\hline
Recall of R-CNN (pix.w.) & 47.50\% &  58.85\% &  48.55\%  & 56.75\%  & 34.99\% &  36.34\% &  53.04\%  & 10.21\%  & 45.21\%  & 13.44\%  & 42.06\%\\
\hline
F-Score of RCNN (pix.w.) & 55.50\% &  61.11\% &  56.69\% &  57.38\%  & 42.84\%  & 40.35\% &  55.09\%  & 16.57\%  & 49.67\%  & 21.93\% &  49.27\%\\
\hline
Precise of Ours (pix.w.) & 83.15\%  &   70.18\%  &  75.97\%  &    89.62\%  &    66.97\%  &    69.96\%  &  61.97\%  &  60.59\%  &  84.27\%  &    86.87\%  &   \textbf{75.52\%}  \\
\hline
Recall of Ours (pix.w.) & 75.41\%  &   70.94\%  &   66.21\%  &   70.77\%  &   75.12\%  &   48.58\%  &   85.41\%  &   37.08\%  &   68.44\%  &   72.66\%  &   \textbf{70.39\%}  \\
\hline
F-Score of Ours (pix.w.) & 79.09\%   &   70.56\%   &   70.75\%   &   79.09\%   &   70.81\%   &   57.34\%   &   71.83\%   &   46.01\%   &   75.54\%   &   79.13\%   &   \textbf{72.87\%}   \\
\hline
\end{tabular}
\end{table*}

Most existing RGBD datasets, e.g. NYU dataset, Washington dataset v2, focus on domestic scene understanding, e.g. distinguishing between tables, chairs, sofa etc., rather than recognition of table-top objects. \kevinupdate{Moreover, these benchmarks are proposed for evaluating fully-supervised methods. In a novel application i.e. industrial object recognition, the large-scale bounding-box annotations are not practical. }

\subsubsection{Dataset}
In order to evaluate our proposed weakly-supervised deep learning approach, we created a novel data-set of industrial objects. Different to most other RGBD recognition challenges (typically involving household or office objects), our application focuses on the major societal problem of robotic decommissioning and cleanup of nuclear waste, which comprises an enormous variety of contaminated objects and materials. In our dataset, there are 217 objects of 10 categories of objects which are common in legacy nuclear waste repositories: plastic bottles, cans, chains, cleaning cloth, gloves, metal objects, plastic pipe, pipe joints, sponges, wood blocks. We randomly split all instances into a training set (147 instances) and a testing set (60 instances), and all testing objects were previously unseen. Our training data is mainly unlabeled RGBD video clips in which a number of training objects are placed on a table. In this experiment, the videos are captured by a Kinect v2 in QHD resolution (540${\times}$960) at 5-6fps. In each video, the camera trajectory covers approximately 180$^{\circ}$ field of view of the objects and the camera poses range from 30$^{\circ}$ to 60$^{\circ}$ above the horizon.

\subsubsection{Baseline Method}
\kevinupdate{Without morderate-scale bounding-box annotations, the existing semi-supervised recognition method \cite{cheng2016semisupervised}, supervised detection \cite{fast-rcnn,yolo,ssd,song2016deep} and recognition methods \cite{eitel2015multimodal,cheng20153dv,cheng2015iccv} are not applicable. Therefore, we use R-CNN \cite{rcnn} as a baseline method, comprising of our 3D-based detector for objectness detection, our pretrained DCNN for feature extraction, and linear SVM for classification.} The R-CNN is trained by using manually labeled objectness proposals. \kevinupdate{And for each proposal image, we adjust the object location as the center (This step is not required in the proposed weakly-supervised approach).} This comparison aims to show the advantage of our proposed weakly-supervised DCNN over fully-supervised approaches such as R-CNN, when very few labeled data are available. 

\subsubsection{Implementation and Running Time}
Our computer has an i7 8-cores CPU and NVIDIA TITAN X GPU (12G). In our implementation, IAI Kinect2 package\footnote{$https://github.com/code-iai/iai_kinect2$} is used to interface with ROS and calibrate RGB and depth cameras. Our DCNN is based on Caffe toolbox\cite{caffe-net}. All of our pipeline is integrated in ROS. The running time of our proposed detection and recognition is 2-3HZ in a QHD point cloud. The detection time is monotonically increasing with the number of 3D points. Moreover, we also devised a lighter DCNN architecture, which can run 3 times faster with only slightly lower performance. Our pipeline can be boosted to 5HZ with point cloud down-sampling and lighter DCNN architecture. In contrast to previous state-of-the-art RGBD object detection methods, 4 seconds per frame was achieved by \cite{lai2014unsupervised} and 16 seconds per frame by \cite{song2016deep}. The performance of our method is an order magnitude greater, and can reasonably be described as near-real-time.

\subsubsection{Training}
23 video clips were captured for training and 3 for testing. In each training video, training objects of a specific category were placed on a table. Each object was captured in different poses and from different viewpoints. Our proposed objectness detection approach generated 163K unlabeled object proposals. We manually labeled 524 examples in total, and trained a binary DCNN-GPC for each category. Statistics of our training data are detailed in Table \ref{tab:result}. Having DCNN-GPC trained by manually labeled examples, the \textit{confidence} (i.e. predictive probabilities) of the 163K unlabeled examples can be estimated by Eq. \ref{eq:confidional_prob}. If the predictive probability of an example is larger than $\tau$, then the prediction is treated as confident and this example is assigned the label of the corresponding category, otherwise it is abandoned. In our implementation, we set $\tau$ as 0.7 for all categories. In this procedure, 56.5K of 163K unlabeled examples are automatically labeled by DCNN-GPC. Then we finetune our multi-modal DCNN using both GP-labeled and manually-labeled examples.

\subsubsection{Evaluation}
As the parameters of our detector are fixed, an ROC curve is not available. Instead, \textit{precision}, \textit{recall} and \textit{f-score} are used for evaluation. Unlike conventional bounding-box-based detection methods, our approach generates boundary-aware (i.e. pixel-wise) detection results. Hence, we evaluate these three measurements both instance-wise and pixel-wise. For evaluation, we first acquire key frames from the four testing videos according to visual odometry. For each video, we uniformly select 10 frames from all key frames. In total, 40 testing frames are obtained. We densely annotated all the objects in these 40 frames (approximately 1000 objects).

\begin{figure*}[thpb]
\centering
\includegraphics[width=0.97\textwidth]{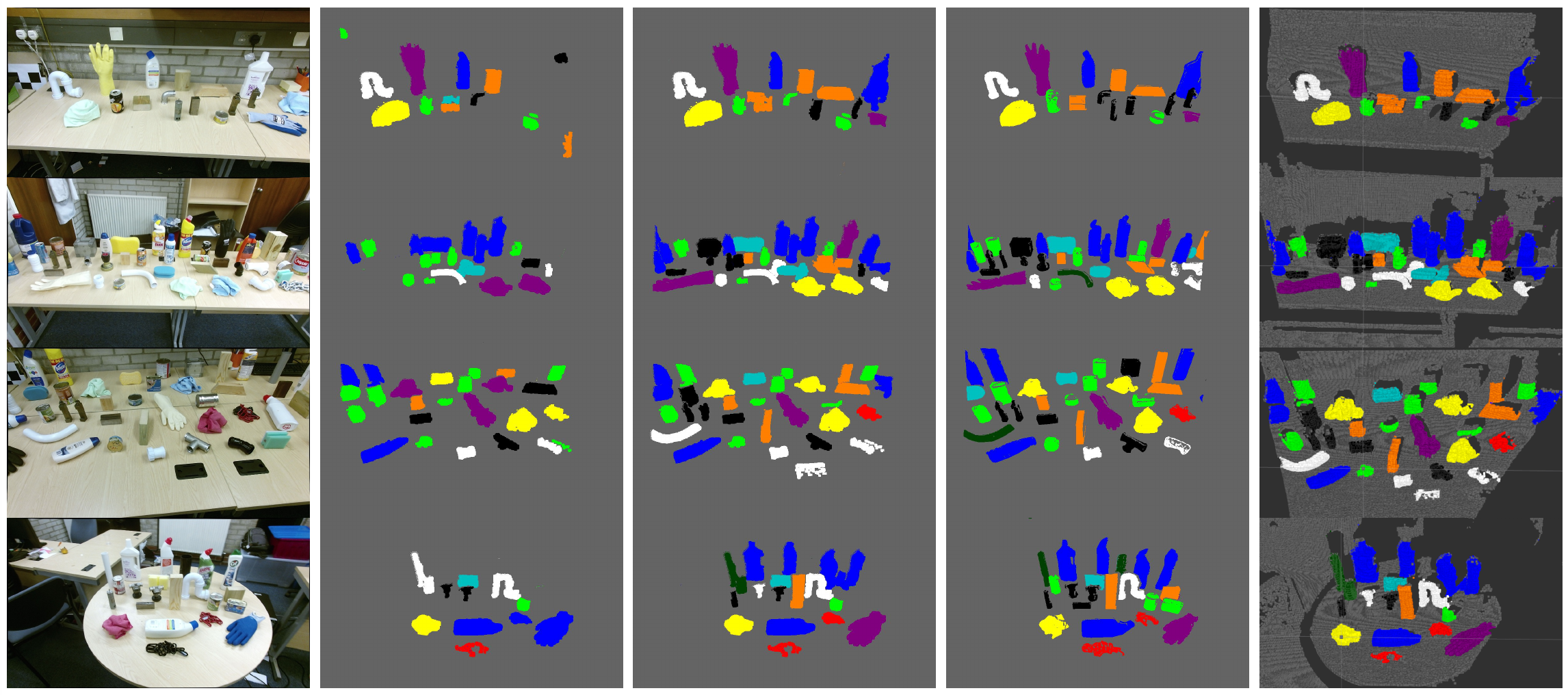}
\label{fig:cnns-gpc}
\caption{\label{fig:qualitative-results}The qualitative results. From left to right: RGB images, 2D semantic maps of R-CNN, 2D semantic maps of our method, the ground truth, and 3D semantic maps of our method.}
\end{figure*}

In the instance-wise evaluation, detections are considered as true or false positives if the overlap area between prediction and ground truth exceeds 50\%. In the pixel-wise evaluation, true or false positives are counted between corresponding pixels. Quantitative results are shown in Table \ref{tab:result} and qualitative results are shown in Fig. \ref{fig:qualitative-results}.

As shown in Table \ref{tab:result}, our approach achieves 80.85\% average precision, 83.53\% recall, 82.17\% f-score in the instance-wise detection test, and 75.52\% average precision, 70.39\% recall, 72.87\% f-score in the pixel-wise detection test. We observe that the difference between instance-wise and pixel-wise performance can be attributed to 3D clustering error, i.e. an object may be segmented as more than one cluster, and the small clusters are ignored because of their small physical dimension. Moreover, the boundary-aware detection is susceptible to point cloud down-sampling, resulting in decreased precision of object boundaries. 

The results suggest that our weakly-supervised DCNN performs substantially better than the fully-supervised R-CNN, when few labeled training examples are available (more than 20\% in F-Score). Compared to R-CNN, our weakly-supervised DCNN is more robust to scale-changes and variance of poses. This is because the moderate-scale automatically labeled data optimizes the DCNN end-to-end.

\section{CONCLUSIONS}
This paper proposed a novel weakly-supervised deep learning approach (DCNN-GPC) for end-to-end learning using minimal annotated data (approximately 50 for each category) by propagating minimal labels to large-scale unlabeled data. We also proposed a novel way to pretrain a DCNN for the depth modality, by using large-scale virtual CAD data, enabling full leveraging of depth data without color-mapping or low-level features. Good adaptation from virtual data to real-world depth data has been demonstrated. Furthermore, a real-time (several frames per second) detection and recognition pipeline has been integrated and demonstrated. Unlike previous methods, bounding-box annotations are not required in training, but boundary-aware detection is achieved. For evaluation, we created a novel industrial object dataset, and demonstrated that DCNNs can be weakly-supervised to effectively solve real-world problems. 





\section*{ACKNOWLEDGMENT}
We thank NVIDIA Corporation for generously donating a high-power GPU on which this work was performed. This work was funded by EU H2020 RoMaNS 645582 and EPSRC EP/M026477/1. Zhao was sponsored by DISTINCTIVE - a university consortium funded by the Research Councils UK Energy programme. Stolkin was sponsored by a Royal Society Industry Fellowship.

\bibliographystyle{IEEEtran}
{\scriptsize
\bibliography{refs_short}}

\addtolength{\textheight}{-12cm}   

\end{document}